# TEI2GO: A Multilingual Approach for Fast Temporal Expression Identification

Hugo Sousa
University of Porto
INESC TEC
Porto, Portugal
hugo.o.sousa@inesctec.pt

Ricardo Campos
University of Beira Interior
INESC TEC
Covilhã, Portugal
ricardo.campos@ubi.pt

Alípio Jorge
University of Porto
INESC TEC
Porto, Portugal
alipio.jorge@inesctec.pt

## ABSTRACT

Temporal expression identification is crucial for understanding texts written in natural language. Although highly effective systems such as HeidelTime exist, their limited runtime performance hampers adoption in large-scale applications and production environments. In this paper, we introduce the TEI2GO models, matching HeidelTime's effectiveness but with significantly improved runtime, supporting six languages, and achieving state-of-the-art results in four of them. To train the TEI2GO models, we used a combination of manually annotated reference corpus and developed "Professor HeidelTime", a comprehensive weakly labeled corpus of news texts annotated with HeidelTime. This corpus comprises a total of 138, 069 documents (over six languages) with 1, 050, 921 temporal expressions, the largest open-source annotated dataset for temporal expression identification to date. By describing how the models were produced, we aim to encourage the research community to further explore, refine, and extend the set of models to additional languages and domains. Code, annotations, and models are openly available for community exploration and use. The models are conveniently on HuggingFace for seamless integration and application.

## 1 INTRODUCTION

Temporal expressions (timexs henceforth) play a pivotal role in written and spoken narratives, providing essential information about the timing of events or actions [5]. Timexs can be broadly categorized into two classes: explicit, such as "26 of May 2023", which can be understood without context, and implicit expressions like "last year", which require additional context for proper interpretation. Understanding timexs is critical for accurate narrative comprehension and is a vital component in the development of temporally aware natural language processing (NLP) systems [29].

The extraction of temporal information from a text begins with the identification of the tokens that refer to timexs, a task known as timex identification (TEI). Due to its foundational importance and widespread implications in NLP applications, TEI has garnered significant interest within the research community. In particular, TEI has been shown to greatly impact downstream tasks such as summarization [15], question answering [38], and, more broadly, in information retrieval [2, 3, 7].

While numerous automatic TEI systems exist [40], each has its limitations. Rule-based models [9, 34], for example, can achieve high effectiveness in specific domains and languages but are labor-intensive to develop, requiring experts to define rules and an annotated corpus to evaluate them. Deep neural models[1, 17], on the other hand, can more easily generalize to new domains and languages, but require substantial amounts of annotated data to achieve a level of effectiveness comparable to rule-based approaches. Nevertheless, the primary bottleneck for existing systems lies in their slow runtime performance, *i.e.*, the amount of time it takes to produce the predictions at inference time. For instance, it took us eight days to annotate roughly 25, 000 documents with HeidelTime (see Section 4 for details about the annotation process). Such runtimes are impractical for most applications and limit the adoption of the systems and, ultimately, the development of temporally aware systems.

To tackle these challenges, we investigate a range of neural architectures and libraries to create TEI models that are effective, scalable, and able to attain production-grade performance. In this paper, we present the TEI2GO models, a set of six models trained for TEI, each tailored to a specific language: English, French, German, Italian, Portuguese, and Spanish. These models stem from the application of the approach we found to be most effective in the context of existing research (Section 2): framing the task as a sequence labeling problem and training the spaCy entity-recognition model on a corpus that combines human and weakly labeled data (Section 3). To train our models, we made use of 15 high-quality data from benchmark datasets and developed a weakly labeled dataset. For the latter, we collected a large collection of news texts, for each of the above-referred languages, and annotated them with HeidelTime. We dubbed this corpus Professor HeidelTime (Section 4). To illustrate the quality of the devised models we present a comprehensive evaluation study, and showcase that the proposed models yield results that are either state-of-the-art (SOTA) or comparable to rule-based systems while retaining the adaptability of neural models and a runtime two orders of magnitude faster than baseline systems (Section 5). In practical terms, this means that a corpus annotation taking one second with our models would require 100 seconds with baseline systems. We conclude by highlighting and discussing future research opportunities (Section 6).

Our main contributions are as follows: (1) we introduce the TEI$_2$GO models, a set of six TEI models that achieve HeidelTime effectiveness with a substantially faster runtime that can be immediately used by the community in any relevant downstream task or application; (2) we make available the annotations of the weakly-labeled dataset produced with HeidelTime – Professor HeidelTime – along with the necessary code to retrieve the documents; (3) we present a comprehensive evaluation on 21 benchmark corpora spanning six languages, where TEI$_2$GO models achieve SOTA results.

## 2 RELATED WORK

Numerous studies have focused on the representation of temporal information in text [5, 20, 31], leading to the emergence of several annotated corpora [4, 11, 14]. The availability of such resources enabled the TempEval shared tasks series [40, 42], which significantly boosted the development of TEI systems. During that period, several highly effective systems emerged, including HeidelTime [34, 36], which is currently the most predominant system for the automatic identification of timexs. Being a rule-based system, HeidelTime was designed to have high effectiveness. However, to attain such effectiveness, an expert is required to create the rules and an annotated corpus is needed to test them, making it challenging to expand into new languages and domains. This motivated Strötgen and Gertz [37] to develop Auto-HeidelTime, which automatically extends the HeidelTime rule set to (practically) every language. Unfortunately, the automatic extension resulted in a significant decrease in the systems' effectiveness. Although other rule-based methods exist [9, 13, 43], they are less effective than HeidelTime.

Since the emergence of large language models [12, 41], most of the NLP benchmark leaderboards have been dominated by them. In the realm of TEI, these pre-trained models have demonstrated adaptability while maintaining moderate effectiveness [1, 8, 10, 17, 18]. The scarcity of annotated data (even for English) has led to a reliance on techniques that compile corpora from multiple languages. For instance, Lange et. al. [17] was able to outperform Auto-HeidelTime in four out-of-domain languages by using adversarial training to align the embeddings spaces across languages. In comparison to HeidelTime, though, their model performed poorly. Another noteworthy work is XLTime [8], which employed multi-task learning and translation to achieve SOTA results in two benchmark corpus (Portuguese and Basque).

Weak supervision [23, 27], another alternative for handling the lack of annotated data, involves training models on corpora labeled programmatically. This idea was first used in the TEI task in the TempEval-3 shared task [40] with the TrainT3 corpus. At the time, the weakly labeled corpus proved detrimental to the effectiveness of the systems. However, recently Almasian et al. [1] leveraged a corpus labeled with HeidelTime to stabilize the training of a transformer-based system for TEI in German. Nevertheless, the corpus has never been made public due to privacy concerns. With Professor HeidelTime we intend to bridge this gap by providing a large weakly labeled corpus for TEI.

Despite considerable efforts in developing a more effective TEI system, current systems do not take runtime performance into consideration. Our models are unique in that they achieve SOTA effectiveness while also having a much faster runtime.

## 3 TEI$_2$GO MODELS

Applying pre-trained large language models to NLP tasks has proven highly beneficial in various scenarios, particularly when dealing with limited annotated data [28]. Although this approach has been shown to improve the SOTA results in two benchmark datasets [8], these models tend to be resource-intensive and often require specialized hardware for efficient execution, rendering them unsuitable for addressing the performance limitations of existing TEI systems. After exploring numerous alternatives, we identified the spaCy entity-recognition model as a suitable approach[1]. Three central aspects set this model apart, making it an ideal candidate for TEI. First, the model leverages hash embeddings, which offer greater efficiency in terms of memory usage and runtime while maintaining comparable effectiveness [30]. Second, the core of the model is built on a transition-based parser [16] that employs a multilayer perceptron to determine the appropriate transition for each token. This design choice results in a fast model with linear-time performance, while retaining the adaptability of deep neural models. Lastly, the model is trained using the BILUO scheme, which has been empirically observed to outperform the traditional IOB scheme [26]. These advantages, coupled with spaCy's built-in features such as tokenization and extensive language support, strengthen the case for leveraging the spaCy entity-recognition model in TEI applications.

The idea behind the TEI$_2$GO models is to train this architecture on a combination of manually annotated reference corpus and weakly labeled data. By combining both approaches, we aim to leverage the best of both worlds, that is, high-quality with large volume.

## 4 PROFESSOR HEIDELTIME

To address the challenges posed by the scarcity of annotated data to train neural models, we constructed a weakly-labeled multi lingual dataset[2], which will be used, together with benchmark datasets, to train our TEI$_2$GO models. For the labeling process, we employed HeidelTime, as it has a large language coverage and presents SOTA results in most of them.

The dataset created encompasses six languages, namely English, French, German, Italian, Portuguese, and Spanish. In the case of Italian, English, German, and French, we were able to find large open-source collections of news documents that met our requirements, namely: a license that allows the redistribution of the dataset, having more than one million tokens, and includes not just the text but also the publication date of each news article which is needed for HeidelTime to normalize the temporal expressions. For these four languages, we release the annotation along with the texts.

---
[1]https://spacy.io/api/entityrecognizer
[2]https://github.com/hmosousa/professor_heideltime

For the remaining two languages, Spanish and Portuguese, we were not able to find open-source collections that matched our criteria. To surmount this obstacle, we created scripts to scrape reference news sources from the respective countries. We are actively collaborating with these news sources to make the datasets publicly available. Meanwhile, we've released the annotation and open-sourced the scripts, complete with a comprehensive guide on how to replicate the compilation of the news articles[3], allowing the community to legally recreate the Spanish and Portuguese collections of Professor HeidelTime. More details about the compiled corpus can be found in the code repositories.

*Annotation Process.* The annotation of the documents was made with HeidelTime Python Wrapper py_heideltime[4]. For the annotation protocol, we allocated a computation time budget of eight days per language. Within this time frame, the automatic annotator sequentially processed the collection, annotating each document at the time. Documents containing non-UTF-8 characters, malformed or missing document creation times, or any HTML code were excluded from the annotation process. Moreover, documents where HeidelTime could not identify any timexs were also omitted from the Professor HeidelTime collection. The final number of annotated documents for each language is presented in Table 1.

Table 1: Professor HeidelTime corpus statistics.

|  | # Docs | # Sents | # Tokens | # Timexs |
| --- | --- | --- | --- | --- |
| English | 24,642 | 725,011 | 18,755,616 | 254,803 |
| Portuguese | 24,293 | 129,101 | 5,929,377 | 111,810 |
| Spanish | 33,266 | 410,806 | 21,617,888 | 348,011 |
| Italian | 9,619 | 135,813 | 3,296,898 | 58,823 |
| French | 27,154 | 53,266 | 1,673,053 | 83,431 |
| German | 19,095 | 634,851 | 12,515,410 | 194,043 |
| **All** | **138,069** | **2,088,848** | **63,788,242** | **1,050,921** |

## 5 EXPERIMENTS

*Corpora.* To collect the corpora, we employed tieval [32], a dedicated Python framework tailored for the development and evaluation of temporal models, facilitating the accumulation of a vast set of benchmark corpora pertinent to the languages incorporated in our experiments. Table 2 depicts the 21 corpora considered (15 benchmarks corpus plus Professor HeidelTime), along with their language, domain, and distribution of the number of timexs over train/validation/test set. The split was performed on the document level with a 80/20 partition for the development/test set. Subsequently, the development set was further divided into an 80/20 ratio for training and validation purposes.[5]

---
[3]For the Spanish corpus: https://github.com/hmosousa/elmundo_scraper. For the Portuguese corpus: https://github.com/hmosousa/publico_scraper.
[4]https://pypi.org/project/py-heideltime
[5]We also conducted preliminary experimentation with the TempEval-2 [42] and MeanTime [22] corpora. However, these were excluded from further investigation since we found that they contain several non-annotated timexs.

Table 2: Distribution of the number of timexs for each of the corpus over the train, validation, and test set. The reference corpus for each language is highlighted in boldface. Column "D" stands for domain and can take to values "Nw" and "Na" representing News and Narratives, respectively.

|  |  | D | Train | Validation | Test |
| --- | --- | --- | --- | --- | --- |
| EN | **TempEval-3** [40] | Nw | 1,472 | 338 | 138 |
|  | TCR [24] | Nw | 126 | 29 | 62 |
|  | AncientTimes [33] | Na | 142 | 125 | 39 |
|  | WikiWars [21] | Na | 2,166 | 117 | 357 |
|  | P. HeidelTime | Nw | 165,385 | 18,469 | 46,307 |
| PT | **TimeBank** [11] | Nw | 911 | 171 | 145 |
|  | P. HeidelTime | Nw | 63,135 | 6,977 | 17,404 |
| ES | **TimeBank** [14] | Nw | 939 | 155 | 198 |
|  | TrainT3 [40] | Nw | 821 | 58 | 215 |
|  | AncientTimes [33] | Na | 152 | 39 | 21 |
|  | P. HeidelTime | Nw | 226,393 | 25,242 | 63,110 |
| IT | **EVENTI-NC** [6] | Nw | 299 | 42 | 98 |
|  | AncientTimes [33] | Na | 184 | 37 | 8 |
|  | P. HeidelTime | Nw | 35,351 | 3,897 | 9,956 |
| FR | **TimeBank** [4] | Nw | 329 | 32 | 64 |
|  | AncientTimes [33] | Na | 144 | 129 | 12 |
|  | P. HeidelTime | Nw | 40,415 | 4,572 | 11,290 |
| DE | **KRAUTS** [39] | Nw | 774 | 98 | 218 |
|  | WikiWars [35] | Na | 1,721 | 98 | 398 |
|  | AncientTimes [33] | Na | 101 | 35 | 55 |
|  | P. HeidelTime | Nw | 126,121 | 13,828 | 34,999 |

*Baselines.* As a baseline reference, we use HeidelTime and Spark NLP NER model [19] to compare with TEI2GO. Whereas HeidelTime provides SOTA effectiveness for the languages in the study, Spark NLP provides a reference for using an alternative deep neural architecture. For the Spark NLP baselines, we produced six models that resulted from training the model on the reference corpus, highlighted in bold face in Table 2, of each language. All the models use the same embeddings, the multilingual GloVe embeddings [25], and were trained with the same rationale that is depicted in the paragraph presented below.

*Training.* For each language, we define three different combinations of datasets to train the model, named: *Base, Compilation*, and *All*. The *Base* models are only trained on the reference corpus for each of the languages. This is a useful baseline since the reference corpus is short but of high quality. However, since all the reference corpora are all of the news domain, we also introduce the *Compilation* models, which are trained in all the corpora gathered for each language, except the weakly-labeled datasets. By doing so, we have a way to evaluate how the model adapts to different domains. Finally, we also train the models in all the corpora gathered for each language, including the weakly-labeled datasets – the *All* models.

To keep the experiments reproducible and comparable we pre-defined 26 hyperparameters configurations to be employed in every model/data combination. Each model was trained for a maximum of 30 epochs with early stopping monitoring validation $F_1$ with patience set to three epochs. The model with the highest validation $F_1$ among the 26 created was kept as our final model. The training and evaluation of the models were executed on a machine with an Intel Core i7-10510U CPU with 16GB of RAM. We encourage interested readers to visit the code repository for further information[6].

Table 3: Results of the devised models on the test set of each corpus. Column $F_1$ and $F_{1_R}$ present the strict and relaxed $F_1$ score [40], respectively. The "Time" column presents the average time it took to compute the predictions for one sentence in milliseconds. In bold are highlighted the best $F_1$ score for each corpus.

| | | Baseline | | | | | | TEI₂GO | | | | | | | | |
| | | HeidelTime | | | SparkNLP | | | Base | | | Compilation | | | All | | |
| | | $F_1$ | $F_{1_R}$ | Time | $F_1$ | $F_{1_R}$ | Time | $F_1$ | $F_{1_R}$ | Time | $F_1$ | $F_{1_R}$ | Time | $F_1$ | $F_{1_R}$ | Time |
|---|---|---|---|---|---|---|---|---|---|---|---|---|---|---|---|---|
| EN | TempEval-3 | 81.8 | 90.7 | 465.1 | 74.9 | 87.5 | 140.6 | 74.6 | 87.7 | 2.8 | 81.2 | 87.8 | 3.8 | **82.4** | 90.6 | 3.8 |
| | TCR | 74.0 | 86.6 | 280.0 | 61.5 | 79.4 | 122.0 | 73.0 | 87.3 | 2.0 | 68.6 | 86.3 | 4.0 | **77.2** | 89.8 | 4.0 |
| | AncientTimes | **89.2** | 91.9 | 150.0 | 35.5 | 53.3 | 154.5 | 11.8 | 35.3 | 4.5 | 68.4 | 84.2 | 4.5 | 66.7 | 74.2 | 4.5 |
| | WikiWars | 83.6 | 91.4 | 62.9 | 50.1 | 74.3 | 92.7 | 50.7 | 78.0 | 2.0 | **90.8** | 96.7 | 2.6 | 84.7 | 91.9 | 2.6 |
| | P. HeidelTime | 100.0 | 100.0 | 276.6 | 57.6 | 70.1 | 134.5 | 69.5 | 82.6 | 2.4 | 71.8 | 81.3 | 3.1 | **98.7** | 99.1 | 3.0 |
| PT | TimeBank | 72.1 | 81.8 | 463.2 | 80.8 | 89.0 | 114.5 | **83.6** | 86.5 | 6.0 | - | - | - | 77.4 | 82.0 | 3.4 |
| | P. HeidelTime | 100.0 | 100.0 | 335.8 | 50.6 | 72.7 | 154.0 | 52.5 | 72.5 | 6.9 | - | - | - | **97.8** | 98.5 | 3.6 |
| ES | TimeBank | **85.6** | 89.1 | 510.3 | 81.6 | 90.4 | 99.3 | 76.6 | 87.1 | 3.4 | 69.3 | 85.6 | 4.8 | **85.6** | 89.1 | 2.8 |
| | TrainT3 | 82.5 | 88.7 | 496.1 | 89.0 | 93.7 | 109.7 | **90.5** | 94.3 | 3.9 | 75.0 | 94.0 | 5.2 | 84.0 | 90.5 | 3.9 |
| | AncientTimes | **78.0** | 92.7 | 130.0 | 15.8 | 75.7 | 75.0 | 15.8 | 75.7 | 5.0 | 51.4 | 76.5 | 5.0 | 38.9 | 75.7 | 0.0 |
| | P. HeidelTime | 100.0 | 100.0 | 418.1 | 68.4 | 82.6 | 249.6 | 66.8 | 82.6 | 4.4 | 60.7 | 80.4 | 7.5 | **97.0** | 98.3 | 3.9 |
| IT | EVENTI-NC | 81.4 | 93.9 | 571.4 | 63.6 | 83.3 | 88.9 | 52.9 | 75.0 | 6.3 | 79.8 | 89.7 | 3.2 | **85.6** | 96.4 | 3.2 |
| | AncientTimes | 36.4 | 54.5 | 86.5 | 28.4 | 49.7 | 159.5 | 36.4 | 54.5 | 2.7 | **82.4** | 82.4 | 0.0 | 46.2 | 61.5 | 0.0 |
| | P. HeidelTime | 100.0 | 100.0 | 364.0 | 52.3 | 72.2 | 84.6 | 52.3 | 66.9 | 5.8 | 52.1 | 76.3 | 2.9 | **98.1** | 98.7 | 2.9 |
| FR | TimeBank | **87.1** | 93.5 | 930.8 | 78.7 | 89.3 | 92.3 | 84.4 | 90.6 | 1.9 | 82.3 | 92.7 | 3.8 | 82.2 | 89.9 | 3.8 |
| | AncientTimes | 87.0 | 95.7 | 192.3 | 10.5 | 31.6 | 76.9 | 0.0 | 22.2 | 0.0 | **88.0** | 96.0 | 0.0 | 44.4 | 66.7 | 0.0 |
| | P. HeidelTime | 100.0 | 100.0 | 297.8 | 70.6 | 82.1 | 106.9 | 68.9 | 79.6 | 2.2 | 71.3 | 85.2 | 2.6 | **98.2** | 98.8 | 2.6 |
| DE | Krauts | **77.7** | 82.7 | 675.6 | 60.6 | 77.8 | 163.7 | 75.9 | 82.2 | 4.4 | 67.1 | 79.2 | 3.7 | 70.5 | 82.9 | 5.2 |
| | WikiWars | **87.3** | 91.9 | 37.3 | 49.3 | 86.5 | 60.1 | 50.4 | 74.0 | 1.5 | 61.5 | 87.7 | 1.3 | 67.6 | 93.6 | 1.7 |
| | AncientTimes | **75.3** | 79.6 | 96.9 | 30.0 | 64.0 | 143.8 | 33.3 | 64.4 | 3.1 | 73.6 | 83.9 | 3.1 | 72.9 | 83.3 | 3.1 |
| | P. HeidelTime | 100.0 | 100.0 | 211.7 | 54.7 | 69.3 | 117.2 | 59.5 | 70.2 | 2.2 | 60.2 | 74.3 | 2.0 | **91.8** | 96.7 | 2.7 |

**Results.** Referring to Table 3, **TEI₂GO** models attain SOTA $F_1$ results in the reference corpus across four of the six evaluated languages (TempEval-3 for English, TimeBank for Portuguese and Spanish, and EVENTI-NC for Italian) while having a two orders of magnitude faster runtime than HeidelTime and SparkNLP (which can be observed by comparing the "Time" column of the **Baseline** and **TEI₂GO** systems). In three languages – English, Spanish, and Italian – the *All* model achieved SOTA results, illustrating the potential advantages of incorporating a weakly labeled corpus in certain scenarios. In contrast, for Portuguese, the introduction of the weakly labeled corpus negatively affected the model's effectiveness, since the SOTA $F_1$ was achieved by the *Base* model. On the remaining two languages – French and German – our models seem to struggle to find the boundaries of timexs, as they achieve comparable values of $F_{1_R}$ to HeidelTime, but are not able to compete in terms of strict $F_1$. Finally, by looking at the effectiveness of the *All* models on the Professor HeidelTime corpus, one can conclude that the proposed approach can distill knowledge from HeidelTime, as it is able to achieve a strict $F_1$ score above 91 per cent in all languages.

As this approach mitigates the bottleneck of current TEI systems, we believe it could foster the emergence of applications of timexs in downstream tasks. To allow easy access and exploration of the models, we published the six TEI₂GO models that have the highest $F_1$ score on the reference corpus of each language on HuggingFace[7].

## 6 CONCLUSION & FUTURE RESEARCH

In this paper, we introduce a set of six models for fast temporal expression identification, dubbed TEI₂GO models. Through a comprehensive evaluation, we demonstrated that the TEI₂GO models are capable of achieving SOTA effectiveness while ensuring production-level performance, which has been a limitation of previous systems. Also highlighted in this manuscript is the fact that the methodology applies to low-resource languages, and can distill knowledge from HeidelTime. Note, however, that since HeidelTime also normalizes the timexs (in addition to its identification), the presented approach is not a full substitute for HeidelTime. For that, further research is required.

## ACKNOWLEDGMENTS

This work is financed by National Funds through the Portuguese funding agency, FCT - Fundação para a Ciência e a Tecnologia, within project LA/P/0063/2020.

---

[6]https://github.com/hmosousa/tei2go

[7]https://huggingface.co/hugosousa

## REFERENCES

[1] Satya Almasian, Dennis Aumiller, and Michael Gertz. 2022. Time for some German? Pre-Training a Transformer-based Temporal Tagger for German. In *Text2Story@ECIR*, Vol. 3117. CEUR-WS, Online.

[2] Omar Alonso, Jannik Strötgen, Ricardo Baeza-Yates, and Michael Gertz. 2011. Temporal information retrieval: Challenges and opportunities. In *CEUR Workshop Proceedings*, Vol. 707. CEUR, Hyderabad, India, 1–8.

[3] Klaus Berberich, Srikanta Bedathur, Omar Alonso, and Gerhard Weikum. 2010. A Language Modeling Approach for Temporal Information Needs. In *Advances in Information Retrieval*. Springer Berlin Heidelberg, Berlin, Heidelberg, 13–25. https://doi.org/10.1007/978-3-642-12275-0_5

[4] André Bittar, Pascal Amsili, Pascal Denis, and Laurence Danlos. 2011. French TimeBank: An ISO-TimeML Annotated Reference Corpus. In *Proceedings of the 49th Annual Meeting of the Association for Computational Linguistics: Human Language Technologies*. Association for Computational Linguistics, Portland, Oregon, USA, 130–134. https://aclanthology.org/P11-2023

[5] Branimir Boguraev and Rie Kubota Ando. 2005. TimeML-Compliant Text Analysis for Temporal Reasoning. In *Proceedings of the 19th International Joint Conference on Artificial Intelligence (IJCAI'05)*. Morgan Kaufmann Publishers Inc., San Francisco, CA, USA, 997–1003.

[6] Alice Bracchi, Tommaso Caselli, and Irina Prodanof. 2016. Enriching the Ita-TimeBank with Narrative Containers. In *Proceedings of the Third Italian Conference on Computational Linguistics*. Accademia University Press, Napoli, 83–88. https://doi.org/10.4000/books.aaccademia.1732

[7] Ricardo Campos, Gaël Dias, Alípio M. Jorge, and Adam Jatowt. 2014. Survey of Temporal Information Retrieval and Related Applications. *Comput. Surveys* 47, 2 (1 2014), 1–41. https://doi.org/10.1145/2619088

[8] Yuwei Cao, William Groves, Tanay Kumar Saha, Joel Tetreault, Alejandro Jaimes, Hao Peng, and Philip Yu. 2022. XLTime: A Cross-Lingual Knowledge Transfer Framework for Temporal Expression Extraction. In *Findings of the Association for Computational Linguistics: NAACL 2022*. Association for Computational Linguistics, Stroudsburg, PA, USA, 1931–1942. https://doi.org/10.18653/v1/2022.findings-naacl.148

[9] Angel X Chang and Christopher Manning. 2012. SUTime: A library for recognizing and normalizing time expressions. In *Proceedings of the Eight International Conference on Language Resources and Evaluation (LREC'12)*, Nicoletta Calzolari (Conference Chair), Khalid Choukri, Thierry Declerck, Mehmet Uğur Doğan, Bente Maegaard, Joseph Mariani, Asuncion Moreno, Jan Odijk, and Stelios Piperidis (Eds.). European Language Resources Association (ELRA), Istanbul, Turkey.

[10] Sanxing Chen, Guoxin Wang, and Börje F Karlsson. 2019. *Exploring Word Representations on Time Expression Recognition*. Technical Report MSR-TR-2019-46. Microsoft Research. https://www.microsoft.com/en-us/research/publication/exploring-word-representations-on-time-expression-recognition/

[11] Francisco Costa and António Branco. 2012. TimeBankPT: A TimeML Annotated Corpus of Portuguese. In *Proceedings of the Eighth International Conference on Language Resources and Evaluation (LREC'12)*. European Language Resources Association (ELRA), Istanbul, Turkey, 3727–3734. http://www.lrec-conf.org/proceedings/lrec2012/pdf/246_Paper.pdf

[12] Jacob Devlin, Ming-Wei Chang, Kenton Lee, and Kristina Toutanova. 2019. BERT: Pre-training of Deep Bidirectional Transformers for Language Understanding. In *Proceedings of the 2019 Conference of the North*. Association for Computational Linguistics, Stroudsburg, PA, USA, 4171–4186. https://doi.org/10.18653/v1/N19-1423

[13] Wentao Ding, Guanji Gao, Linfeng Shi, and Yuzhong Qu. 2019. A Pattern-Based Approach to Recognizing Time Expressions. *Proceedings of the AAAI Conference on Artificial Intelligence* 33, 01 (7 2019), 6335–6342. https://doi.org/10.1609/aaai.v33i01.33016335

[14] Marta Guerrero Nieto, Roser Saurí, and Miguel-Angel Bernabé-Poveda. 2011. ModeS TimeBank: A modern Spanish TimeBank Corpus. *Revista de la Sociedad Española del Procesamiento del Lenguaje Natural* 47 (6 2011), 259–267.

[15] Philip Hausner, Dennis Aumiller, and Michael Gertz. 2020. Time-Centric Exploration of Court Documents. In *Proceedings of Text2Story - Third Workshop on Narrative Extraction From Texts co-located with 42nd European Conference on Information Retrieval,*. CEUR Workshop Proceedings, 31–37.

[16] Matthew Honnibal and Mark Johnson. 2015. An Improved Non-monotonic Transition System for Dependency Parsing. In *Proceedings of the 2015 Conference on Empirical Methods in Natural Language Processing*. Association for Computational Linguistics, Lisbon, Portugal, 1373–1378. https://doi.org/10.18653/v1/D15-1162

[17] Lukas Lange, Anastasiia Iurshina, Heike Adel, and Jannik Strötgen. 2020. Adversarial Alignment of Multilingual Models for Extracting Temporal Expressions from Text. In *Proceedings of the 5th Workshop on Representation Learning for NLP*. Association for Computational Linguistics, Online, 103–109. https://doi.org/10.18653/v1/2020.repl4nlp-1.14

[18] Egoitz Laparra, Dongfang Xu, and Steven Bethard. 2018. From Characters to Time Intervals: New Paradigms for Evaluation and Neural Parsing of Time Normalizations. *Transactions of the Association for Computational Linguistics* 6 (2018), 343–356.

[19] Xuezhe Ma and Eduard Hovy. 2016. End-to-end Sequence Labeling via Bi-directional LSTM-CNNs-CRF. In *Proceedings of the 54th Annual Meeting of the Association for Computational Linguistics (Volume 1: Long Papers)*. Association for Computational Linguistics, Berlin, Germany, 1064–1074. https://doi.org/10.18653/v1/P16-1101

[20] Inderjeet Mani and George Wilson. 2000. Robust Temporal Processing of News. In *Proceedings of the 38th Annual Meeting of the Association for Computational Linguistics*. Association for Computational Linguistics, Hong Kong, 69–76. https://doi.org/10.3115/1075218.1075228

[21] Pawel Mazur and Robert Dale. 2010. WikiWars: A New Corpus for Research on Temporal Expressions. In *Proceedings of the 2010 Conference on Empirical Methods in Natural Language Processing*. Association for Computational Linguistics, Cambridge, MA, 913–922. https://aclanthology.org/D10-1089

[22] Anne-Lyse Minard, Manuela Speranza, Ruben Urizar, Begoña Altuna, Marieke van Erp, Anneleen Schoen, and Chantal van Son. 2016. MEANTIME, the NewsReader Multilingual Event and Time Corpus. In *Proceedings of the Tenth International Conference on Language Resources and Evaluation*. European Language Resources Association (ELRA), Portorož, Slovenia, 4417–4422.

[23] Mike Mintz, Steven Bills, Rion Snow, and Daniel Jurafsky. 2009. Distant supervision for relation extraction without labeled data. In *Proceedings of the Joint Conference of the 47th Annual Meeting of the ACL and the 4th International Joint Conference on Natural Language Processing of the AFNLP*. Association for Computational Linguistics, Suntec, Singapore, 1003–1011. https://aclanthology.org/P09-1113

[24] Qiang Ning, Zhili Feng, Hao Wu, and Dan Roth. 2018. Joint Reasoning for Temporal and Causal Relations. In *Proceedings of the 56th Annual Meeting of the Association for Computational Linguistics (Volume 1: Long Papers)*. Association for Computational Linguistics, Stroudsburg, PA, USA, 2278–2288. https://doi.org/10.18653/v1/P18-1212

[25] Jeffrey Pennington, Richard Socher, and Christopher Manning. 2014. GloVe: Global Vectors for Word Representation. In *Proceedings of the 2014 Conference on Empirical Methods in Natural Language Processing (EMNLP)*. Association for Computational Linguistics, Doha, Qatar, 1532–1543. https://doi.org/10.3115/v1/D14-1162

[26] Lev Ratinov and Dan Roth. 2009. Design Challenges and Misconceptions in Named Entity Recognition. In *Proceedings of the Thirteenth Conference on Computational Natural Language Learning (CoNLL-2009)*. Association for Computational Linguistics, Boulder, Colorado, 147–155. https://aclanthology.org/W09-1119

[27] Livy Real, Alexandre Rademaker, Fabricio Chalub, and Valeria De Paiva. 2018. Towards Temporal Reasoning in Portuguese. In *6th Workshop on Linked Data in Linguistics: Towards Linguistic Data Science*.

[28] Sebastian Ruder, Matthew E Peters, Swabha Swayamdipta, and Thomas Wolf. 2019. Transfer Learning in Natural Language Processing. In *Proceedings of the 2019 Conference of the North American Chapter of the Association for Computational Linguistics: Tutorials*. Association for Computational Linguistics, Minneapolis, Minnesota, 15–18. https://doi.org/10.18653/v1/N19-5004

[29] Brenda Santana, Ricardo Campos, Evelin Amorim, Alípio Jorge, Purificação Silvano, and Sérgio Nunes. 2023. A Survey on Narrative Extraction from Textual Data. *Artificial Intelligence Review* (1 2023). https://doi.org/10.1007/s10462-022-10338-7

[30] Joan Serrà and Alexandros Karatzoglou. 2017. Getting Deep Recommenders Fit: Bloom Embeddings for Sparse Binary Input/Output Networks. In *Proceedings of the Eleventh ACM Conference on Recommender Systems (RecSys '17)*. Association for Computing Machinery, New York, NY, USA, 279–287. https://doi.org/10.1145/3109859.3109876

[31] Andrea Setzer. 2001. *Temporal information in newswire articles : an annotation scheme and corpus study*. Ph. D. Dissertation.

[32] Hugo Sousa, Ricardo Campos, and Alípio Mário Jorge. 2023. tieval: An Evaluation Framework for Temporal Information Extraction Systems. In *Proceedings of the 46th International ACM SIGIR Conference on Research and Development in Information Retrieval*. ACM, New York, NY, USA, 2871–2879. https://doi.org/10.1145/3539618.3591892

[33] Jannik Strötgen, Thomas Bögel, Julian Zell, Ayser Armiti, Tran Van Canh, and Michael Gertz. 2014. Extending HeidelTime for Temporal Expressions Referring to Historic Dates. In *Proceedings of the Ninth International Conference on Language Resources and Evaluation (LREC'14)*. European Language Resources Association (ELRA), Reykjavik, Iceland, 2390–2397. http://www.lrec-conf.org/proceedings/lrec2014/pdf/849_Paper.pdf

[34] Jannik Strötgen and Michael Gertz. 2010. HeidelTime: High Quality Rule-Based Extraction and Normalization of Temporal Expressions. In *Proceedings of the 5th International Workshop on Semantic Evaluation*. Association for Computational Linguistics, Uppsala, Sweden, 321–324. https://aclanthology.org/S10-1071

[35] Jannik Strötgen and Michael Gertz. 2011. WikiWarsDE : A German Corpus of Narratives Annotated with Temporal Expressions. In *Proceedings of the Conference of the German Society for Computational Linguistics and Language Technology*. Hamburger Zentrum für Sprachkorpora, Hamburg, Germany, 129–134.

[36] Jannik Strötgen and Michael Gertz. 2013. Multilingual and Cross-Domain Temporal Tagging. *Language Resources and Evaluation* 47, 2 (6 2013), 269–298.

https://doi.org/10.1007/s10579-012-9179-y
[37] Jannik Strötgen and Michael Gertz. 2015. A Baseline Temporal Tagger for all Languages. In *Proceedings of the 2015 Conference on Empirical Methods in Natural Language Processing*. Association for Computational Linguistics, Stroudsburg, PA, USA, 541–547. https://doi.org/10.18653/v1/D15-1063
[38] Jannik Strötgen and Michael Gertz. 2016. *Domain-Sensitive Temporal Tagging*. Synthesis Lectures on Human Language Technologies, Vol. 9. Springer International Publishing, Cham. https://doi.org/10.1007/978-3-031-02163-3
[39] Jannik Strötgen, Anne-Lyse Minard, Lukas Lange, Manuela Speranza, and Bernardo Magnini. 2018. KRAUTS: A German Temporally Annotated News Corpus. In *Proceedings of the Eleventh International Conference on Language Resources and Evaluation (LREC 2018)*, Nicoletta Calzolari, Khalid Choukri, Christopher Cieri, Thierry Declerck, Koiti Hasida, Hitoshi Isahara, Bente Maegaard, Joseph Mariani, Asuncion Moreno, Jan Odijk, Stelios Piperidis, and Takenobu Tokunaga (Eds.). European Language Resources Association (ELRA), Miyazaki, Japan, 536–540. https://aclanthology.org/L18-1085
[40] Naushad UzZaman, Hector Llorens, Leon Derczynski, James Allen, Marc Verhagen, and James Pustejovsky. 2013. Semeval-2013 task 1: Tempeval-3: Evaluating time expressions, events, and temporal relations. In *Second Joint Conference on Lexical and Computational Semantics (* SEM), Volume 2: Proceedings of the Seventh International Workshop on Semantic Evaluation (SemEval 2013)*. Association for Computational Linguistics, Atlanta, Georgia, USA, 1–9.
[41] Ashish Vaswani, Noam Shazeer, Niki Parmar, Jakob Uszkoreit, Llion Jones, Aidan N. Gomez, Lukasz Kaiser, and Illia Polosukhin. 2017. Attention Is All You Need. In *Advances in Neural Information Processing Systems*, Vol. abs/1706.03762. Neural Information Processing Systems Foundation, Inc. (NeurIPS), Long Beach, California, USA, 5998–6008. http://arxiv.org/abs/1706.03762
[42] Marc Verhagen, Roser Sauri, Tommaso Caselli, and James Pustejovsky. 2010. SemEval-2010 Task 13: TempEval-2. In *Proceedings of the 5th international workshop on semantic evaluation*. Association for Computational Linguistics, Uppsala, Sweden, 57–62.
[43] Xiaoshi Zhong, Aixin Sun, and Erik Cambria. 2017. Time Expression Analysis and Recognition Using Syntactic Token Types and General Heuristic Rules. In *Proceedings of the 55th Annual Meeting of the Association for Computational Linguistics (Volume 1: Long Papers)*. Association for Computational Linguistics, Vancouver, Canada, 420–429. https://doi.org/10.18653/v1/P17-1039